\documentclass[10pt, conference, compsocconf]{IEEEtran}
\usepackage[table,xcdraw]{xcolor}

\usepackage{amssymb} 

\usepackage{listings}

\definecolor{codegreen}{rgb}{0,0.6,0}
\definecolor{codegray}{rgb}{0.5,0.5,0.5}
\definecolor{codepurple}{rgb}{0.58,0,0.82}
\definecolor{backcolour}{rgb}{0.95,0.95,0.92}

\lstdefinestyle{mystyle}{
    backgroundcolor=\color{backcolour},   
    commentstyle=\color{codegreen},
    keywordstyle=\color{magenta},
    numberstyle=\tiny\color{codegray},
    stringstyle=\color{codepurple},
    basicstyle=\ttfamily\footnotesize,
    breakatwhitespace=false,         
    breaklines=true,                 
    captionpos=b,                    
    keepspaces=true,                 
    numbers=left,                    
    numbersep=5pt,                  
    showspaces=false,                
    showstringspaces=false,
    showtabs=false,                  
    tabsize=2
}
\usepackage{fancyhdr}

\ifCLASSINFOpdf
  \usepackage[pdftex]{graphicx}
  \usepackage{circuitikz}
  \usepackage{tikz}
\else
\fi
\usepackage[cmex10]{amsmath}
\usepackage{algorithm}
\usepackage{algpseudocode}

\usepackage[caption=false,font=footnotesize]{subfig}
\newcommand{\myheader}{
    Presented in the 21st Conference on Robots and Vision (CRV 2024) Workshop
}

\fancypagestyle{plain}{%
    \fancyhf{}
    \fancyhead[L]{\myheader}
}

\begin{document}

%
\title{Phasor-Driven Acceleration for FFT-based CNNs}


\author{\IEEEauthorblockN{Eduardo Reis}
\IEEEauthorblockA{\textit{Dept. of Electrical and Computer Eng.} \\
\textit{Lakehead University}\\
Thunder Bay, Canada \\
edreis@lakeheadu.ca}
\and
\IEEEauthorblockN{Thangarajah Akilan}
\IEEEauthorblockA{\textit{Dept. of Software Eng.} \\
\textit{Lakehead University}\\
Thunder Bay, Canada \\
takilan@lakeheadu.ca }
\and
\IEEEauthorblockN{Mohammed Khalid}
\IEEEauthorblockA{\textit{Dept. of Electrical and Computer Eng.} \\
\textit{University of Windsor}\\
Windsor, Canada \\
mkhalid@uwindsor.ca}
}


%


\maketitle

\thispagestyle{plain}  


%
%

\begin{abstract}

Recent research in deep learning (DL) has investigated the use of the Fast Fourier Transform (FFT) to accelerate the computations involved in Convolutional Neural Networks (CNNs) by replacing spatial convolution with element-wise multiplications on the spectral domain.
These approaches mainly rely on the FFT to reduce the number of operations, which can be further decreased by adopting the Real-Valued FFT.
In this paper, we propose using the phasor form---a polar representation of complex numbers, as a more efficient alternative to the traditional approach.
The experimental results, evaluated on the CIFAR-10, demonstrate that our method achieves superior speed improvements of up to a factor of
1.376 (average of 1.316) during training and up to 1.390 (average of 1.321)  during inference when compared to the traditional rectangular form employed in modern CNN architectures.
Similarly, when evaluated on the CIFAR-100, our method achieves superior speed improvements of up to a factor of 1.375 (average of 1.299) during training and up to 1.387 (average of 1.300) during inference.
Most importantly, given the modular aspect of our approach, the proposed method can be applied to any existing convolution-based DL model without design changes.

\end{abstract}
  

\IEEEpeerreviewmaketitle


\section{Introduction}

{
CNNs have become the cornerstone of the recent advancements in computer vision applications, leading the way for models with \textit{better-than-human} performance across various tasks, viz. image classification, image enhancement, and object detection, recognition, and tracking. 
However, training the state-of-the-art Deep Convolutional Neural Networks (DCNNs) requires large-scale datasets, such as ImageNet~\cite{ImageNet2014}, to be processed many times over, thus making the computation speed of the model a critical aspect.

Traditionally, DCNNs are known to take considerable amounts of time to be trained, 
even when powered by Graphical Processing Units (GPUs):
{AlexNet}~\cite{AlexNet2012} takes 5-6 days when trained on \texttt{2x NVIDIA GTX 580 3GB} GPUs;
{VGG}~\cite{Vgg2015} variants takes 2-3 weeks on \texttt{4x NVIDIA Titan Black} GPUs;
{Inception-V3}~\cite{InceptionV3_2016} and {ResNet-50}~\cite{ResNet2016} take,
respectively, 25 and 18 days on a single \texttt{NVIDIA Quadro P4000}
GPU, as found in~\cite{Zhu2018}. 
Hence, recent studies show that with abundant computational power, the training time can be reduced by orders of several magnitude:
{ResNet-50} takes 20 minutes to be trained on \texttt{2048x Intel Xeon Phi 7250}~\cite{You2018}.
The training time of {ResNet-50} is lowered to 122 seconds on \texttt{3456x NVIDIA Tesla V100}~\cite{Mikami2019};
and further reduced to 74.7 seconds on \texttt{2048x NVIDIA Tesla V100}~\cite{Yamazaki2019}.

Although the computational time required by DCNN models can be massively reduced by using clusters with a high number of processing units, such approaches do not focus on accelerating computations by reducing the number of operations on an algorithmic level.
Researching techniques to lower the computational requirements of DCNNs is still a highly demanding task due to two unique reasons:
(a) not every application has access to abundant computational power since GPUs are still an expansive resource and edge applications might be restricted to a couple of CPU cores; (b) such techniques are platform agnostic; hence, they can also be incorporated into applications that have many processing units to further reduce their computational time and energy consumption footprint.
}

{
Several strategies have been proposed to accelerate DCNNs in literature. For instance, the work in \cite{Zhang2019} proposes a taxonomy of CNN acceleration methods, which categorizes the techniques present in the literature in three main levels: {structure level}, {algorithm level}, and {implementation level}.
As highlighted in that taxonomy, three approaches are used to obtain efficient convolution operations at the algorithm level, as listed below. 
\begin{itemize}
\item \emph{im2col-based} algorithms, which are based on
the GEneral Matrix Multiplication (GEMM) function of the
Basic Linear Algebra Subprograms (BLAS) library~\cite{BLAS};
\item {Winograd}'s algorithms~\cite{Lavin2016}, which are capable of computing minimal arithmetic complexity for convolutions of small kernels and mini-batches sizes; and
\item FFT-based algorithms~\cite{Mathieu2014, Vasilache2015}, which are based on the convolution theorem of the Fourier domain.
\end{itemize}
These three methods are well-established; they, and their variations, serve as canonical approaches within the \texttt{NVIDIA cuDNN} library~\cite{CuDNN}.
The vast majority of CNN applications use \texttt{cuDNN} indirectly through frameworks, such as PyTorch~\cite{PyTorch2019}, which by defaults causes the \texttt{cuDNN} to benchmark the convolution algorithms and select the fastest.
For instance, PyTorch's benchmark process relies on \texttt{cuDNN} functions like \texttt{cudnnFindConvolutionForwardAlgorithm} and \texttt{cudnnFindConvolutionBackwardDataAlgorithm}.
Hence, these three algorithms are of extreme importance, and any improvements made to them could be integrated into \texttt{cuDNN} and seamlessly improve all applications relying on them.
}

In this paper, we propose a method to speed FFT-based convolution using phasors to reduce the number of operations to perform the spectral domain convolution.

  %
  The remainder of this paper is organized as follows.
  Section~\ref{sec:related-work} provides a summary of related research.
  Section~\ref{sec:methodology} presents the proposed method to accelerate Fourier-based CNN using phasor.
  Section~\ref{sec:experimental_results} compares the performance of the proposed approach with existing methods.
  Section~\ref{sec:discussion} discusses the limitations of the proposed approach.
  Finally, Section~\ref{sec:conclusion} concludes the paper with directions for further research.

\section{Related Work}\label{sec:related-work}

{
The use of FFT to accelerate CNNs was first proposed by Mathieu~\textit{et al.}~\cite{Mathieu2014}.
The authors introduce a straightforward method that significantly speeds up both training and inference stages.
This speedup is accomplished by replacing the convolution implementation with a point-wise product in the Fourier domain.
Such gain is possible since the input is processed using mini-batches, which enables the reuse of the kernel spectral representation over each input sample.
Thus, the cost of this approach in Floating-point Multiplications (FLOPs) is approximated by \eqref{eq:flops-baseline}.
\begin{equation} \label{eq:flops-baseline}
2 C N^2 \log_2 N [Bf_1 + Bf_2 + f_2 f_1] + 4 B f_2 f_1 N^2,
\end{equation}
where $2 C N^2 \log_2 N$ is the cost for the 2-D
FFT of a given image of size $N$ by $N$;
$B$ is the mini-batch size; 
$f_1$ is the number of input feature maps; and 
$f_2$ is the number of output feature maps.
There is a hidden cost of $C$ in the FFT that is associated
with cropping and discarding some coefficients of the
output.
%
Additionally, the authors in \cite{Mathieu2014} suggest 
taking advantage of the Hermitian symmetry of the FFT for real-valued inputs.
Hence, the memory and computation costs can be reduced by
a factor of nearly half.
Such advantage can be obtained by simply swapping the FFT with the Real-valued FFT (RFFT), which yields $N$ by 
$ \left \lfloor \frac{N}{2} \right \rfloor + 1$ 
instead of $N$ by $N$ frequency components.
}

{
Similarly, the authors in \cite{Vasilache2015} introduced two new implementations for FFT-based convolutions using GPUs.
Both approaches have their performance profiled and extensively examined against the standard convolution implementation of the NVIDIA \texttt{cuDNN} library.
The first implementation is based on NVIDIA's \texttt{cuFFT} and \texttt{cuBLAS} libraries, achieving $1.4\times$--$14.5\times$ speedup over the NVIDIA \texttt{cuDNN }implementation.
%
%
The second implementation, named \texttt{fbfft}, available in 
the Facebook CUDA library~\cite{FacebookCudaLibrary},
provides a significant speedup of over $1.5\times$ when compared to their first implementation.
Though the \texttt{fbfft} can yield superior performance,
it performs poorly when using batches with sizes less than
$8$ and over $64$.
}

{
Highlander and Rodriguez in \cite{Highlander2015} also proposed an FFT-based convolution.
As highlighted by the authors, such convolution methods have a bottleneck on the FFT cost, which is estimated to $\mathcal{O}(N^2 \log_2 N)$ FLOPs. 
To mitigate this, they used \emph{Overlap-and-Add} technique,
reducing the computational complexity to $\mathcal{O}(N^2 \log_2 K)$.  
This considerably increases the efficiency when $N$ is far larger than $K$, which is the case in CNNs.
The results show their method reduces computational time
up to $16.3\times$ of the traditional convolution
implementation for a kernel of size $8$ by $8$ and an
image of size $224$ by $224$.
}

{
Abtahi~\textit{et al.}~\cite{Abtahi2018}, on the other hand,  argued that large CNNs are computationally intensive.
Thus, deploying them in embedded platforms requires very
optimized implementations.
That said, the authors propose a series of experiments to
find the most suitable convolution implementation for 
each specific embedded hardware for the ResNet-20~\cite{ResNet2016} architecture.
%
%
The investigated convolution implementations are the traditional convolution, the FFT-based convolution, and the FFT \emph{Overlap-and-Add} convolution.
The embedded platforms used for the experiments are the {Power-Efficient Nano-Clusters (PENCs)} many-core architecture, the {ARM Cortex A53} CPU, 
the {NVIDIA Jetson TX1} GPU, and the {SPARTCNet} accelerator on the {Zynq 7020} FPGA.
}

{
Lin and Yao~\cite{Lin2019} pointed out that decomposing the convolutions in the spatial domain, as in \cite{Lavin2016}, is more suitable for small kernels while decomposing in the Fourier domain would be more suitable for large inputs.
Considering these aspects, they proposed a novel decomposition strategy in the Fourier domain to accelerate convolution for large inputs with small kernels. 
The algorithm called \texttt{tFFT} implements tile-sized transformations in the Fourier domain. 
They evaluate the performance of \texttt{tFFT} by implementing it on a set of state-of-the-art CNNs.
Their results show that the \texttt{tFFT} reduces the average arithmetic complexity by over $2.64$ compared to the conventional FFT-based convolution algorithms at batch sizes from $1$ to $128$.
}

In summary, most of the existing works on FFT-based CNNs focus on reducing the cost of the FFT operation for smaller kernel sizes. 
This is reasonable because the FFT operation accounts for most of the processing cost of FFT-based CNNs.
%
However, the existing literature overlooks alternative approaches to accelerate FFT-based CNNs, concentrating solely on optimizing the FFT operation without exploring other potential methods.
For example, the spectral domain element-wise convolutions, performed after the FFT operation over the input and kernels, also have relevant processing costs.
To the best of our knowledge, no work has investigated alternative ways to reduce the complexity of the spectral domain element-wise convolutions as proposed in this paper.
\section{Methodology}\label{sec:methodology}

{  
Most DNN frameworks, such as PyTorch, implement the forward operation of convolution layers as a cross-correlation instead of convolution.
Each convolution layer depends on three convolution operations that are part of the feed-forward and back-propagation phases of the network:
the one convolution for the feed-forward is given in \eqref{eq:forward-conv}; and two convolutions for the back-propagation are given in \eqref{eq:backward-conv-input} and~\eqref{eq:backward-conv-kernel}.
\begin{equation} \label{eq:forward-conv}
y_{f_2} = \sum_{f_1} x_{f_1} \star w_{f_2 f_1}, 
\end{equation}
\begin{equation} \label{eq:backward-conv-input}
\frac{\partial L}{\partial x_{f_1}} = \frac{\partial L}{\partial y_{f_2}} \ast w^{T}_{f_2 f_1},
\end{equation}
\begin{equation} \label{eq:backward-conv-kernel}
\frac{\partial L}{\partial w_{f_2 f_1}} = \frac{\partial L}{\partial y_{f_2}} \star x_{f_1},
\end{equation}
where we have the convolution operator $\ast$, the cross-correlation operator $\star$, the loss $L$, 
the input feature map $x_{f_1}$ of size $N$ by $N$,
the kernel $w_{f_2 f_1}$ of size $K$ by $K$,
and the output $y_{f_2}$.

Our method replaces each of these convolutions,
which are traditionally performed in the spatial domain,
by their equivalent Fourier-domain operation using a new phasor approach.
Fig.~\ref{fig:methodology} depicts the steps involved in the proposed method.
These steps are:

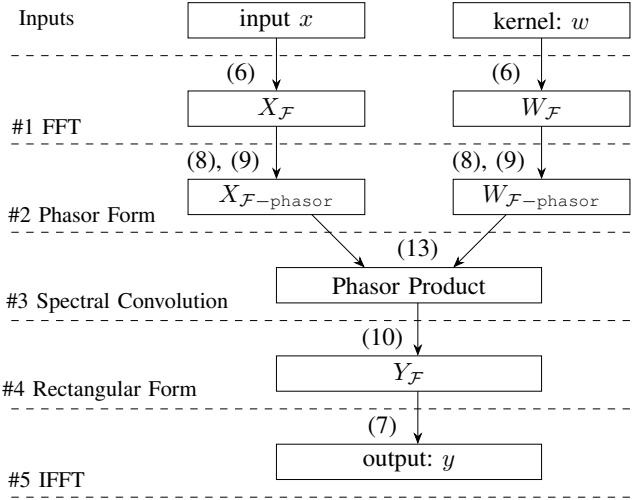
\begin{figure}[!tp]
\centering
\resizebox{1\columnwidth}{!}{%
\begin{circuitikz}
\tikzstyle{every node}=[font=\normalsize]
\draw  (7.5,16) rectangle  node {\normalsize input $x$} (10,15.5);
\draw  (11.25,16) rectangle  node {\normalsize kernel: $w$} (13.75,15.5);
\draw  (7.5,14.75) rectangle  node {\normalsize $X_\mathcal{F}$} (10,14.25);
\draw  (11.25,14.75) rectangle  node {\normalsize $W_\mathcal{F}$} (13.75,14.25);
\draw [->, >=Stealth] (8.75,15.5) -- (8.75,14.75);
\draw [->, >=Stealth] (12.5,15.5) -- (12.5,14.75);
\node [font=\normalsize] at (8.25,15) {\eqref{eq:dft}};
\node [font=\normalsize] at (12,15) {\eqref{eq:dft}};
\draw [dashed] (5,14) -- (13.75,14);
\node [font=\small] at (5.5,14.25) {\#1 FFT};
\draw  (7.5,13.5) rectangle  node {\normalsize $X_{\mathcal{F}-\texttt{phasor}}$} (10,13);
\draw  (13.75,13) rectangle  node {\normalsize $W_{\mathcal{F}-\texttt{phasor}}$} (11.25,13.5);
\draw [->, >=Stealth] (8.75,14.25) -- (8.75,13.5);
\draw [->, >=Stealth] (12.5,14.25) -- (12.5,13.5);
\draw [dashed] (5,15.25) -- (13.75,15.25);
\node [font=\small] at (5.5,15.75) {Inputs};
\node [font=\small] at (6,13) {\#2 Phasor Form};
\draw [dashed] (5,12.75) -- (13.75,12.75);
\draw [dashed] (5,11.5) -- (13.75,11.5);
\draw [dashed] (5,10.25) -- (13.75,10.25);
\draw [dashed] (5,9) -- (13.75,9);
\node [font=\normalsize] at (11.75,13.75) {\eqref{eq:complex-abs}, \eqref{eq:complex-angle}};
\node [font=\normalsize] at (8,13.75) {\eqref{eq:complex-abs}, \eqref{eq:complex-angle}};
\draw  (8.75,12.25) rectangle  node {\normalsize Phasor Product} (12.5,11.75);
\draw  (8.75,11) rectangle  node {\normalsize $Y_\mathcal{F}$} (12.5,10.5);
\draw  (8.75,9.75) rectangle  node {\normalsize output: $y$} (12.5,9.25);
\draw [->, >=Stealth] (9.25,13) -- (10,12.25);
\draw [->, >=Stealth] (12,13) -- (11.25,12.25);
\node [font=\normalsize] at (10.75,12.5) {\eqref{eq:phasor_product}};
\node [font=\small] at (6.5,11.75) {\#3 Spectral Convolution};
\draw [->, >=Stealth] (10.75,11.75) -- (10.75,11);
\node [font=\small] at (6.25,10.5) {\#4 Rectangular Form};
\node [font=\normalsize] at (10.25,11.25) {\eqref{eq:rectangular}};
\node [font=\small] at (5.5,9.25) {\#5 IFFT};
\draw [->, >=Stealth] (10.75,10.5) -- (10.75,9.75);
\node [font=\normalsize] at (10.25,10) {\eqref{eq:idft}};
\end{circuitikz}
}%
\caption{Overview of the proposed method using phasor product to reduce the number of operations between $X_{\mathcal{F}}$ and $W_{\mathcal{F}}$.}
\label{fig:methodology}
\end{figure}

\begin{enumerate}
\item Fourier transform (cf.~\eqref{eq:dft})
\item Phasor conversion (cf.~\eqref{eq:complex-abs}, and \eqref{eq:complex-angle}),
\item Spectral domain product (cf~\eqref{eq:phasor_product}),
\item Rectangular form conversion (cf~\eqref{eq:rectangular}), and
\item Inverse Fourier transform (cf.~\eqref{eq:idft}).
\end{enumerate}

The operations applied in these steps are elaborated in the following subsections.
For simplicity, the notation used here will mainly focus on 1-D signals, but it can be extended to 2-D signals without any loss of generalization.
}
  
{
\subsection{Fourier Transform} \label{fft}

The Discrete Fourier Transform (DFT) is an orthogonal
a transformation that provides the spectral representation.
The DFT of a given input signal, $x[n]$ is described in \eqref{eq:dft},
where $e^{j\theta}$ is derived from Euler's formula,
\eqref{eq:euler}.
Except for a normalizing factor and the direction of the
rotation on the complex plane, the inverse DFT is similar to
the DFT, as shown in \eqref{eq:idft}.
\begin{equation}
e^{j\theta} = \cos(\theta) + j \sin(\theta) \label{eq:euler}  \end{equation}
\begin{equation}
X_{\mathcal{F}}[k] = \sum_{n=0}^{N-1} x[n] e^{-j 2 \pi k n / N}  \label{eq:dft}
\end{equation}
\begin{equation}
  x[n] = \frac{1}{\sqrt{N}} \sum_{n=0}^{N-1} X_{\mathcal{F}}[k] e^{j 2 \pi k n / N} \label{eq:idft}
\end{equation}
FFT algorithms are provided by most deep learning libraries, e.g., \texttt{Pytorch} and \texttt{cuDNN}, 
and they have computational complexity of $\mathcal{O}(N \log (N))$.
When having a real-valued input, $x[n]$,
its FFT, $X_{\mathcal{F}}(k)$, is Hermitian symmetric,
that is, $X_{\mathcal{F}}(-k) = {X_{\mathcal{F}}}^{*}(k)$.
In such cases, only $\left\lfloor N/2 \right\rfloor + 1$ 
elements need to be computed for an input of size $N$.
Since the convolution layer inputs are real numbers, our method uses the RFFT, thus processing only half of the number of frequency components.
}

{
\subsection{Complex Representation}

The FFT outputs a complex number, $X_{\mathcal{F}}[k]$,
for each frequency component $k$.
Traditionally, complex numbers, $z$, are represented in the 
rectangular form, i.e., $z = a + jb$, having $a$ and $b$ as coefficients to represent, respectively, the real and imaginary axes.
In this work, we propose using phasors, the polar form of complex numbers, to represent the FFT transforms of 
the inputs, $X_\mathcal{F}$, and convolution kernel weights, $W_\mathcal{F}$.
The polar form, a.k.a the exponential form, 
represents a complex number, $z$,
as a vector in the complex domain, $\mathbf{z}$,
based on its norm $|\mathbf{z}|$, \eqref{eq:complex-abs},
and angle $\phi$, \eqref{eq:complex-angle}.
Hence, the phasor $z$ can be written in its exponential form as $|\mathbf{z}| e^{j\phi}$, or in its polar form as $|\mathbf{z}|\angle\phi$.
The rectangular form of $z$ is easily obtained using \eqref{eq:rectangular}.
\begin{equation}
|\mathbf{z}| = \sqrt{a^2 + b^2} \label{eq:complex-abs}
\end{equation}
\begin{equation}
\phi = \tan^{-1}{\left ( \frac{b}{a} \right )} \label{eq:complex-angle}
\end{equation}
\begin{equation}
\mathbf{z} 
    = |\mathbf{z}| \cos(\phi) + j|\mathbf{z}| \sin(\phi)   \label{eq:rectangular}
\end{equation}

}

{
\subsection{Spectral Domain Product}

The spatial domain convolution can be computed as a product of
the FFT representation of the input and the filter kernel, \eqref{eq:fft_conv}.
\begin{equation}
    x[n] \ast w[n] = \mathcal{F}^{-1} \left \{ 
        X_{\mathcal{F}}[k] \cdot 
        W_{\mathcal{F}}[k] 
    \right \} \label{eq:fft_conv}
\end{equation}
The traditional approach for computing the spectral domain convolution is based on the rectangular representation.
In this form, the complex multiplication requires $2$ real-valued additions and  $4$ real-valued multiplications, \eqref{eq:rect_product}. 
\begin{equation}
    z_1 z_2 = (a_1 a_2 - b_1 b_2) + j (a_1 b_2+ a_2 b_1) \label{eq:rect_product}
\end{equation}

In this work, we propose using the phasors to multiply the FFT transforms of both inputs and convolution kernel, hence reducing the number of operations to only $1$ addition and $1$ multiplication, as defined in \eqref{eq:phasor_product}.
\begin{equation}
z_1 z_2 = |\mathbf{z}_1| \cdot |\mathbf{z}_2| \ \angle \ \phi_1 + \phi_2  \label{eq:phasor_product}
\end{equation}
}

Using the above computational strategies, our method speeds up the operations in FFT-based CNNs. 
\section{Experimental Analysis} \label{sec:experimental_results}

\subsection{Environment Configuration}

All implementations are entirely developed in \texttt{Python 3.10}, using the \texttt{PyTorch 1.12.1} and \texttt{Torchvision 0.13.1} frameworks, and the experiments are conducted on a machine that has a \texttt{Intel(R) Xeon(R) Gold 6148 CPU @ 2.4 GHz}, \texttt{92 GB} of RAM, and a \texttt{NVIDIA Tesla P40} with \texttt{22919 MiB}.

\subsection{Baseline}

The FFT-based convolution using the RFFT is chosen as the baseline.
This method uses the rectangular form to compute the complex multiplication in the spectral domain.
This approach is based on the works of  \cite{Mathieu2014}, which have extensively demonstrated that CNNs can greatly benefit from FFT-based convolutions to accelerate computation.

\subsection{Implementation Details}

We implement the baseline as a Python class that inherits from \texttt{pytorch.autograd.Function}.
The \texttt{forward} method is implemented according to \eqref{eq:forward-conv}, and the \texttt{backward} method according to \eqref{eq:backward-conv-input}, and \eqref{eq:backward-conv-kernel}.
Then, the method \texttt{conv2d} is implemented.
The PyTorch framework will call this method whenever a model using a convolution layer is built or loaded.
We use this \texttt{conv2d} method to apply our implementation only to the convolution layers that fit the following conditions:
\begin{itemize}
    \item Kernel is size $(K, K)$;
    Image is $(N, N) | N \ge K$;
    \item Padding is $(P, P)$;
    Stride $(S)$ is $(1, 1)$;
    \item Dilation $(D)$ is $(1, 1)$;
    Groups is $1$.
\end{itemize}
When such a condition is not met, the \texttt{conv2d} defaults to the standard PyTorch implementation.
Next, we implement a method \texttt{spectral\_operation}  to calculate the product between the representation of the two signals in the FFT domain.
This method is used in both the \texttt{forward} and the \texttt{backward} methods.
The proposed method inherits the baseline method class, having only to specialize the \texttt{spectral\_operation} method.
The pseudo-code of the baseline implementation and our method are shown, respectively, in Algorithm~\ref{alg:baseline} and Algorithm~\ref{alg:our_method}.
In addition, we implement a Python context manager \texttt{OverrideConv2d} as shown in Listing~\ref{lst:our_ctx_manager}, which enables us to apply either the baseline or the proposed method implementation to any existing PyTorch model by simply wrapping the model execution code with \texttt{OverrideConv2d(<method>):} statements.

\begin{algorithm}[!tp]
\caption{Spectral operation function for the baseline method based on \cite{Mathieu2014}.}
\label{alg:baseline}
\begin{algorithmic}[1]

\Statex
\Function{Conv2dRFFT}{$x$: Tensor, $w$: Tensor} 
    \State DECLARE $a$, $b$, $c$, $d$ \textbf{as} Tensor
    \State $a$, $b \gets \text{real\_part}(x)$, $\text{imag\_part}(x)$
    \State $c$, $d \gets \text{real\_part}(w)$, $\text{imag\_part}(w)$
    \State \textbf{return} $(a \times c - b \times d) + (b \times c + a \times d) \times \textbf{1j}$
\EndFunction
\end{algorithmic}
\end{algorithm}

\begin{algorithm}[!tp]
\caption{Proposed phasor-driven spectral operation.}
\label{alg:our_method}
\begin{algorithmic}[1]
\Function{Conv2dRFFTPhasor}{$x$: Tensor, $w$: Tensor}
    \State DECLARE $a$, $b$, $c$, $d$ \textbf{as} Tensor
    \State $a$, $b \gets \text{abs}(x)$, $\text{angle}(x)$
    \State $c$, $d \gets \text{abs}(w)$, $\text{angle}(w)$
    \State \textbf{return} $a \times c \times \text{exp}((b + d) \times \textbf{1j})$
\EndFunction
\end{algorithmic}
\end{algorithm}

\begin{lstlisting}[float=th,language=Python, caption=The context manager developed in this work., label=lst:our_ctx_manager]
from torch import autograd
from torch.nn import functional as F

class OverrideConv2d(object):
    """Replaces only conv2d operation by the given function."""

    def __init__(self, new_function: autograd.Function):
        assert type(new_function) in [autograd.function.FunctionMeta, type(None)]
        self._fn = F.conv2d
        self._new_fn = new_function

    def __enter__(self):
        F.__dict__[self._fn.__name__] = (
            self._new_fn.conv2d if self._new_fn is not None else self._fn
        )

    def __exit__(self, *args):
        F.__dict__[self._fn.__name__] = self._fn
\end{lstlisting} 

\begin{table*}[!ht]
\centering
\caption{Batch Processing Time Analysis: Our Method Outperforms the Baseline (Based on \cite{Mathieu2014}), for training on CIFAR-10.}
\label{tab:model_batch_profiling_cifar10}
{
\begin{tabular}{|c|c|c|c|c|c|c|}
\hline \hline
Architecture &
  \begin{tabular}[c]{@{}c@{}}Batch Size\end{tabular} &
  Method &
  \begin{tabular}[c]{@{}c@{}}Total Time \\ (sec)\end{tabular} &
  \begin{tabular}[c]{@{}c@{}}Speedup\\ ($T_b/T_m$)\end{tabular} \\ \hline \hline
VGG-16                  
& 4           
& Baseline            
& 13.893          
& 1.000          
\\ \hline
\rowcolor{green!10}
{VGG-16}         
& {4}  
& {Our Method} 
& {11.019} 
& {1.261} 
\\ \hline
DenseNet-121            
& 8           
& Baseline            
& 17.876          
& 1.000          
\\ \hline
\rowcolor{green!10}
{DenseNet-121}   
& {8}  
& {Our Method} 
& {13.476} 
& {1.326} 
\\ \hline
EfficientNetB3          
& 16          
& Baseline            
& 20.337          
& 1.000          
\\ \hline
\rowcolor{green!10}
{EfficientNetB3} 
& {16} 
& {Our Method} 
& {14.967} 
& {1.359} 
\\ \hline
Inception-V3             
& 16          
& Baseline            
& 40.967          
& 1.000          
\\ \hline
\rowcolor{green!10}
{Inception-V3}    
& {16} 
& {Our Method} 
& {29.222} 
& {1.402} 
\\ \hline
AlexNet                 
& 64          
& Baseline            
& 5.978           
& 1.000          
\\ \hline
\rowcolor{green!10}
{AlexNet}        
& {64} 
& {Our Method} 
& {4.433}  
& {1.349} 
\\ \hline
ResNet-18               
& 64          
& Baseline            
& 19.676          
& 1.000          
\\ \hline
\rowcolor{green!10}
{ResNet-18}      
& {64} 
& {Our Method} 
& {14.310} 
& {1.375} 
\\ \hline \hline
\end{tabular}
}
\end{table*}
\begin{table*}[ht]
\centering
\caption{Batch Processing Time Analysis: Our Method Outperforms the Baseline (Based on \cite{Mathieu2014}), for training on CIFAR-100.}
\label{tab:model_batch_profiling_cifar100}
{
\begin{tabular}{|c|c|c|c|c|c|c|}
\hline \hline
Architecture &
  \begin{tabular}[c]{@{}c@{}}Batch Size\end{tabular} &
  Method &
  \begin{tabular}[c]{@{}c@{}}Total Time \\ (sec)\end{tabular} &
  \begin{tabular}[c]{@{}c@{}}Speedup\\ ($T_b/T_m$)\end{tabular} \\ \hline \hline
VGG-16         & 4  & Baseline   & 13.904 & 1.000 \\ \hline \rowcolor{green!10}
VGG-16         & 4  & Our Method & 11.027 & 1.261 \\ \hline
DenseNet-121   & 4  & Baseline   & 9.809  & 1.000 \\ \hline \rowcolor{green!10}
DenseNet-121   & 4  & Our Method & 7.946  & 1.234 \\ \hline
EfficientNetB3 & 8  & Baseline   & 10.856 & 1.000 \\ \hline \rowcolor{green!10}
EfficientNetB3 & 8  & Our Method & 8.394  & 1.293 \\ \hline
Inception-V3   & 8  & Baseline   & 22.427 & 1.000 \\ \hline \rowcolor{green!10}
Inception-V3   & 8  & Our Method & 17.320 & 1.295 \\ \hline
AlexNet        & 64 & Baseline   & 5.922  & 1.000 \\ \hline \rowcolor{green!10}
AlexNet        & 64 & Our Method & 4.409  & 1.343 \\ \hline
ResNet-18      & 64 & Baseline   & 19.615 & 1.000 \\ \hline \rowcolor{green!10}
ResNet-18      & 64 & Our Method & 14.303 & 1.371 \\ \hline \hline
\end{tabular}
}
\end{table*}

\subsection{Problem Domain, Dataset, and DCNN Architectures}

The proposed model is tested in a Transfer Learning (TL) application, in which several DCNN models pre-trained on the ImageNet dataset are fine-tuned to create image classifiers for the CIFAR-10 and CIFAR-100 datasets.
%
The DCNN models are implemented as provided by the \texttt{TorchVision} library, in which their \texttt{DEFAULT} weights are used as an initial stage of the fine-tuning process.
We adopt the following network architectures:
AlexNet,
DenseNet-121,
EfficientNetB3,
Inception-V3,
ResNet-18, and
VGG-16 with batch-normalization.

\subsection{Batch Execution Speedup}

We use the \texttt{torch.profiler.profile} tool to measure the total processing time.
The profile schedule parameters are set as follows:
\texttt{skip\_first=0}, 
\texttt{wait=4}, 
\texttt{warmup=4}, 
\texttt{active=4}, and
\texttt{repeat=1}.
The total time is averaged by the number of executions the profiling was active, then summarized in Table~\ref{tab:model_batch_profiling_cifar10}, which shows our method yields gains of from $1.261 \times$ to $1.371 \times$ in Speedup time for all six DCNN architectures on the CIFAR-10.
Our method speedup gains of from $1.234 \times$ to $1.371 \times$ on the
CIFAR-100, as shown in Table~\ref{tab:model_batch_profiling_cifar100}.

\subsection{Transfer Learning (TL) Speedup}

\begin{figure}[ht]

\subfloat[VGG-16]{\includegraphics[width=\columnwidth]
{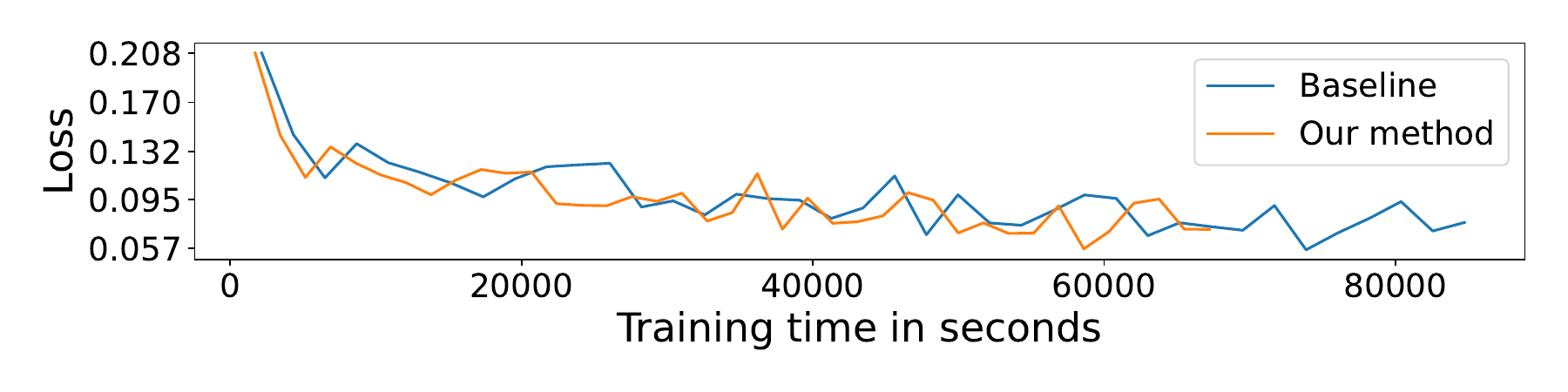}} \\
\subfloat[DenseNet-121]{\includegraphics[width=\columnwidth]
{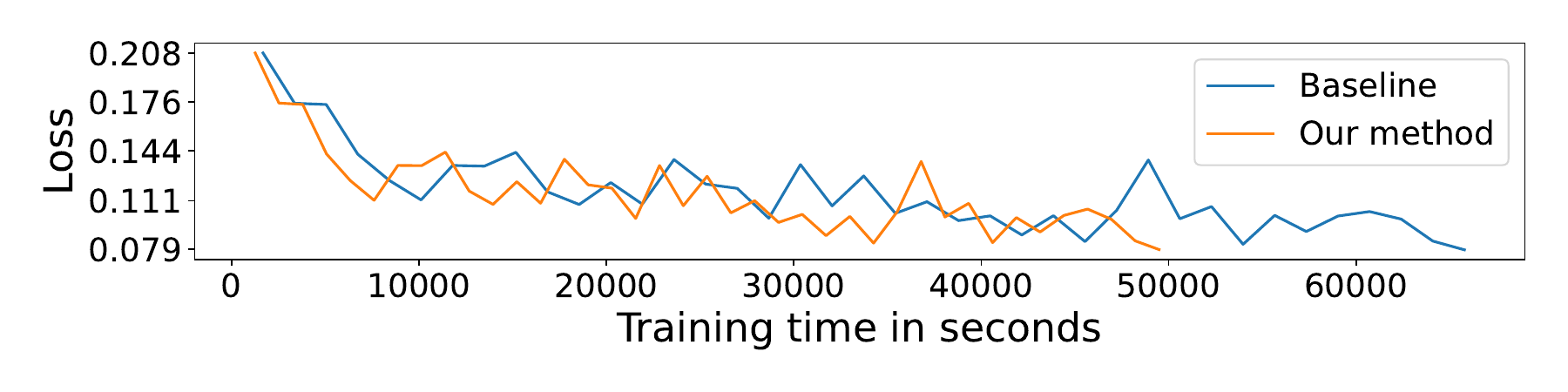}} \\
\subfloat[AlexNet]{\includegraphics[width=\columnwidth]
{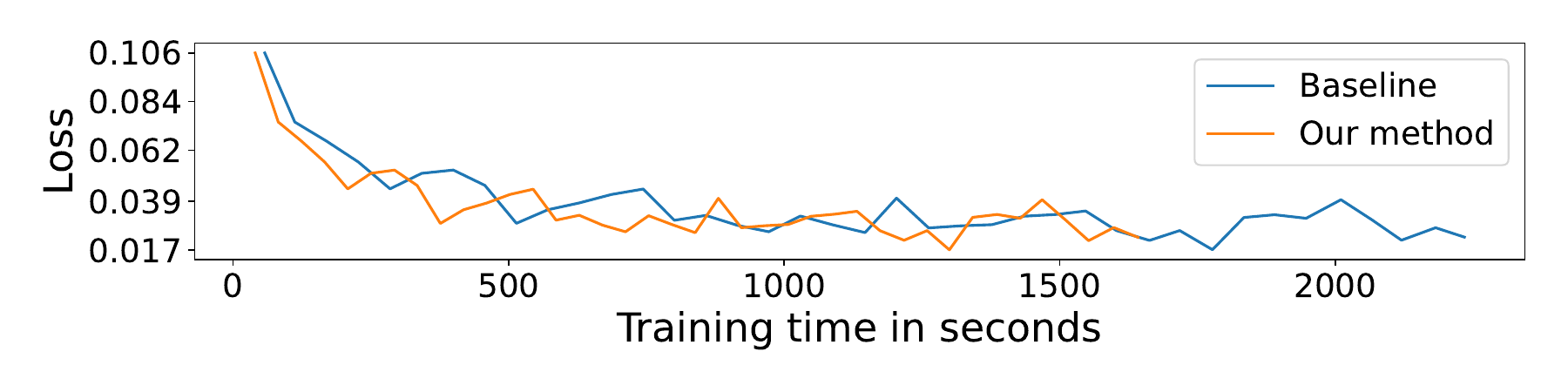}}  \\
\subfloat[EfficientNetB3]{\includegraphics[width=\columnwidth]
{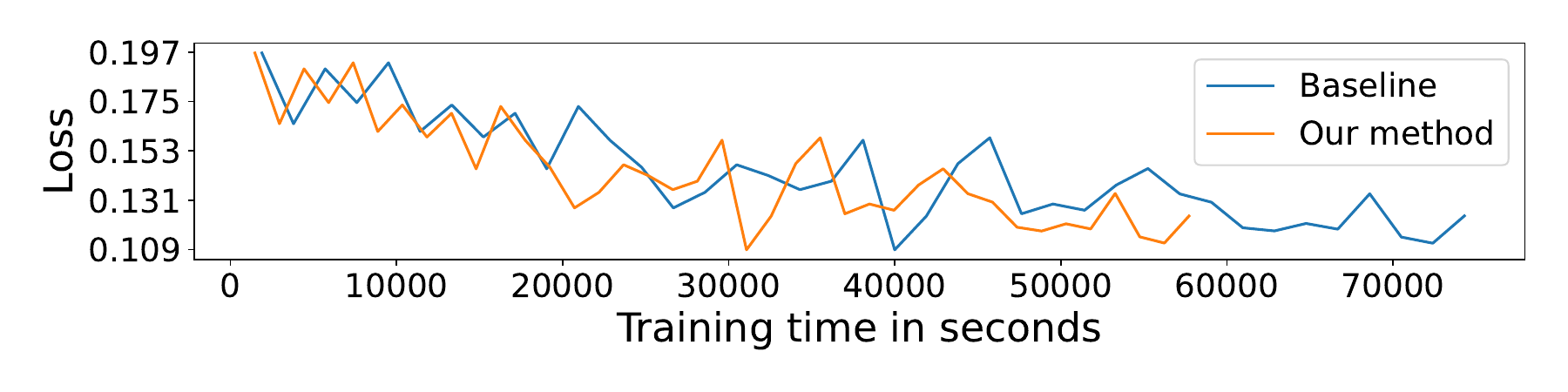}}\\
\subfloat[Inception-V3]{\includegraphics[width=\columnwidth]
{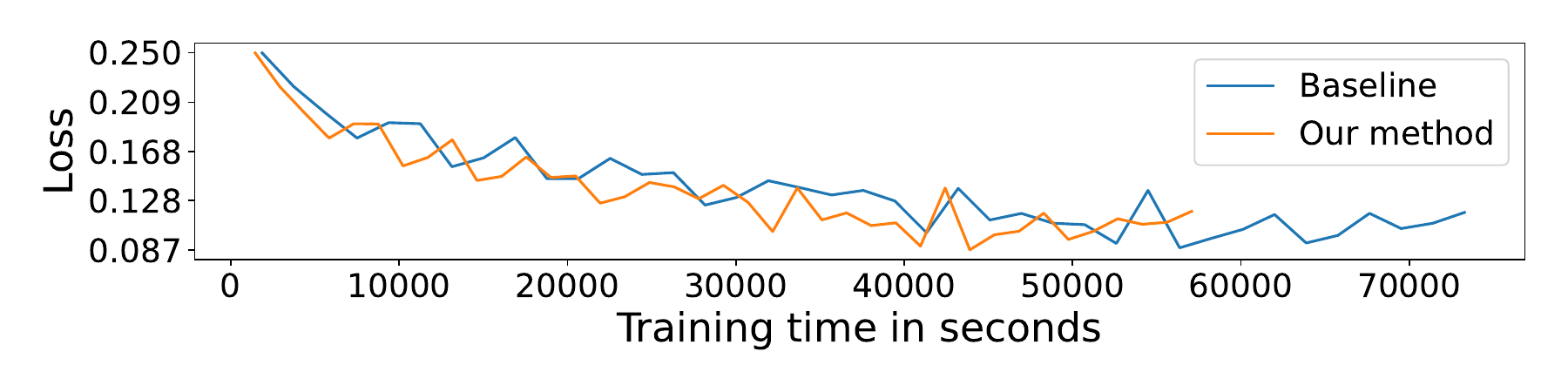}} \\
\subfloat[ResNet-18]{\includegraphics[width=\columnwidth]
{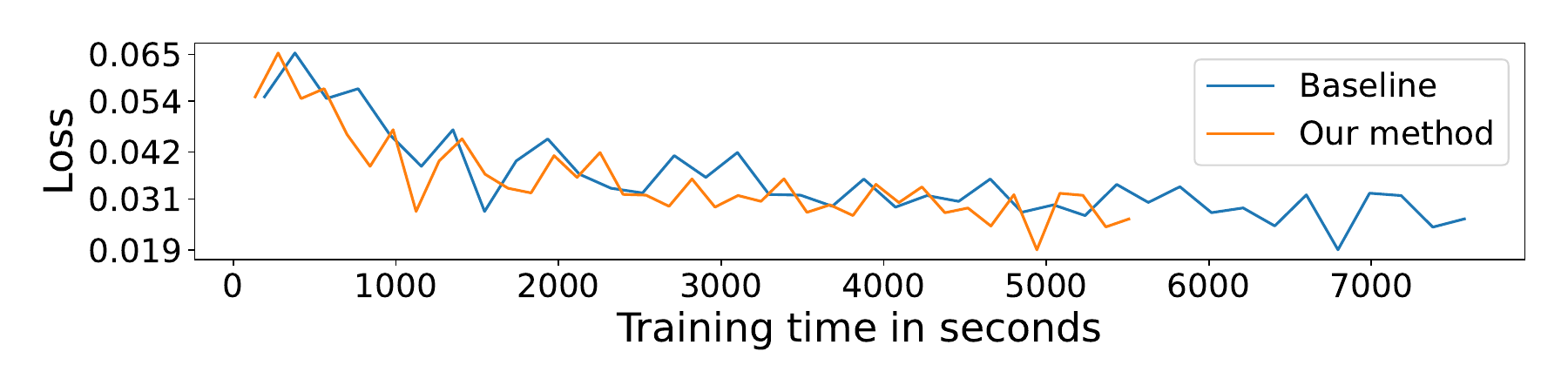}}
\caption{Training loss comparison of the proposed model w.r.t. the baseline,
based on \cite{Mathieu2014}, for six different networks on CIFAR-10.
}
\label{fig:training_loss_cifar10}
\end{figure}

\begin{figure}[ht]

\subfloat[VGG-16]{\includegraphics[width=\columnwidth]
{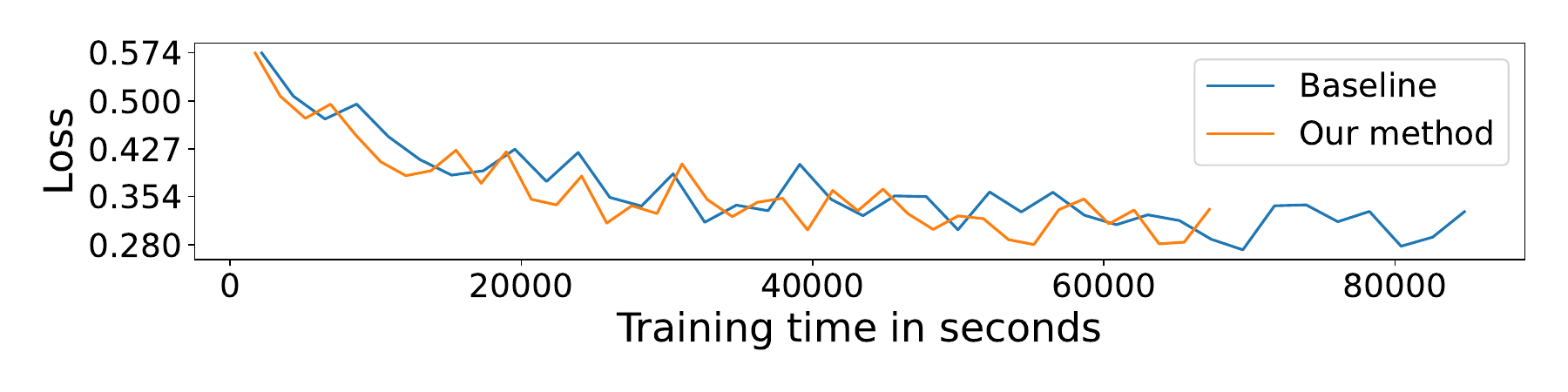}} \\
\subfloat[DenseNet-121]{\includegraphics[width=\columnwidth]
{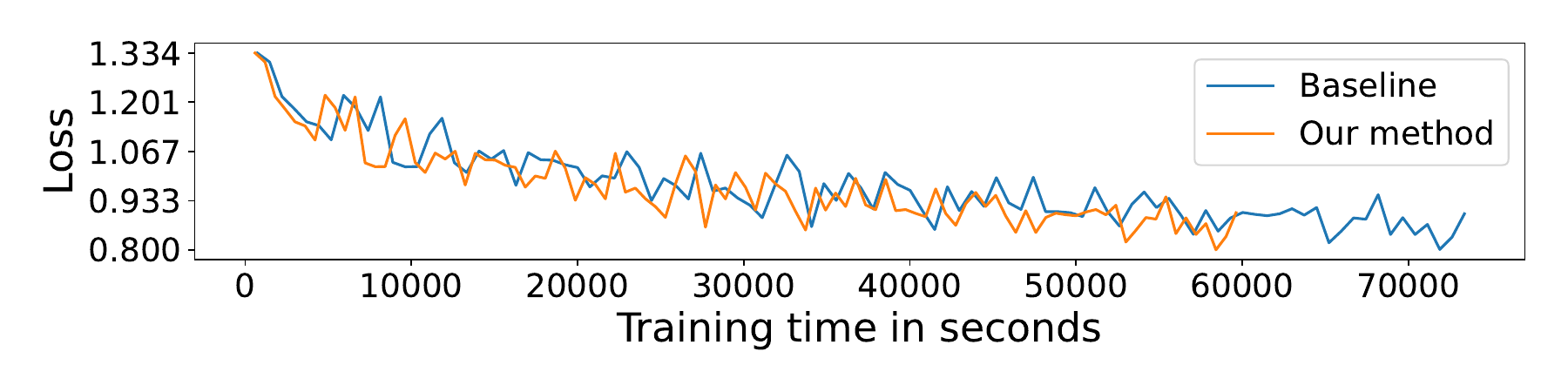}} \\
\subfloat[AlexNet]{\includegraphics[width=\columnwidth]
{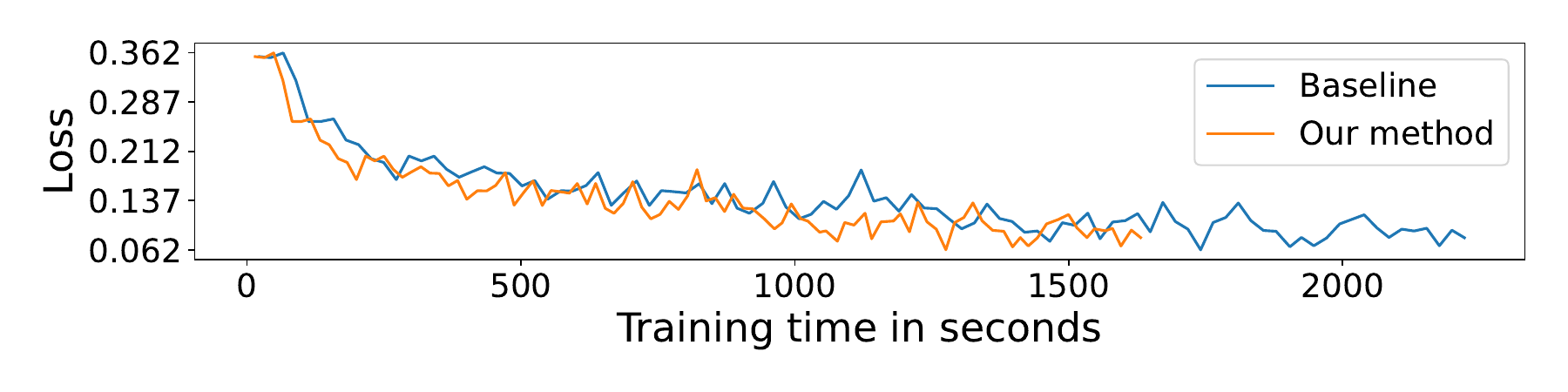}}  \\
\subfloat[EfficientNetB3]{\includegraphics[width=\columnwidth]
{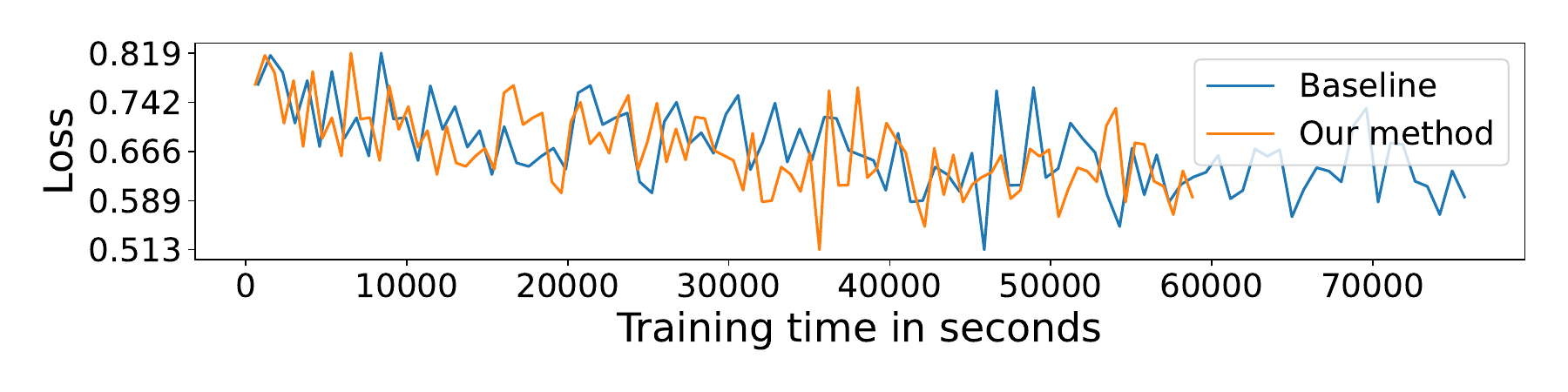}}\\
\subfloat[Inception-V3]{\includegraphics[width=\columnwidth]
{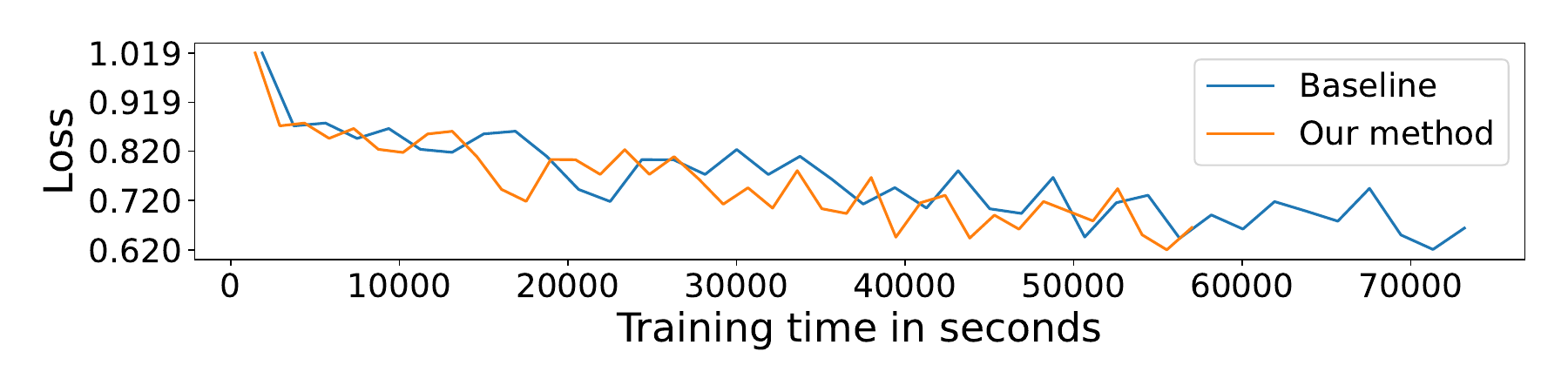}} \\
\subfloat[ResNet-18]{\includegraphics[width=\columnwidth]
{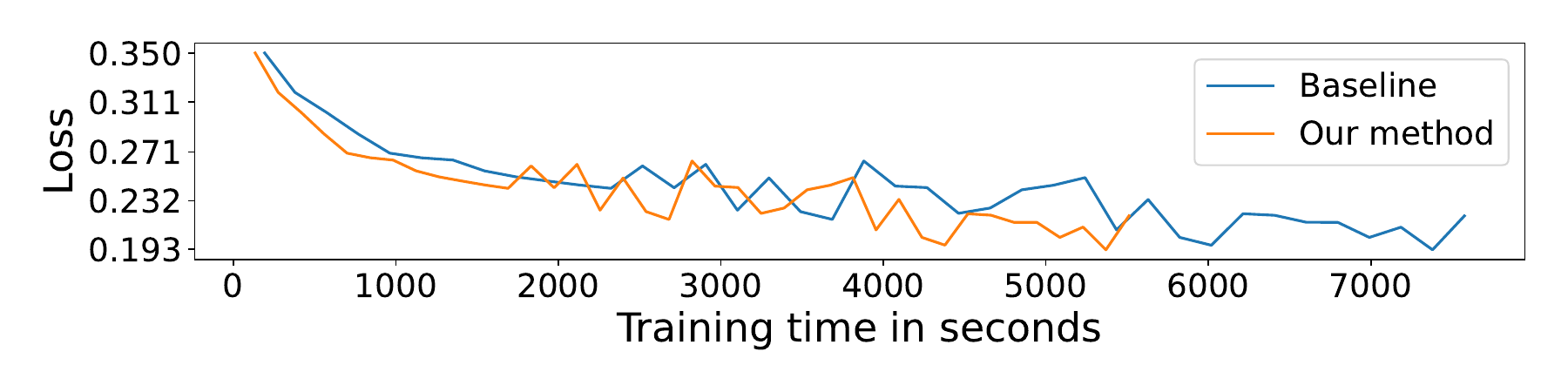}}
\caption{Training loss comparison of the proposed model w.r.t. the baseline,
based on \cite{Mathieu2014}, for six different networks on CIFAR-100.
}
\label{fig:training_loss_cifar100}
\end{figure}

Each DCNN model used in this work is adapted by changing the top layer of the network to make a 10-way and 100-way classification to address the problem domain considered in this study.
Then, the network is trained using \texttt{torch.optim.Adam} optimizer with \texttt{lr=1e-5}, for $2$ epochs having only the classifier layer unfrozen, followed by another $2$ epochs with the entire network unfrozen.
Finally, the network is trained for an epoch, during which we compare our proposed method to the baseline approach.
Table~\ref{tab:model_training_time_cifar10} summarizes the speedup comparison with the baseline, which shows our method yields speedup gains above $1.258 \times$ (and $1.321\times$ on average) for training all six DCNN architectures on the CIFAR-10, while still maintaining the model's learning performance.
Similarly, Table~\ref{tab:model_training_time_cifar100} summarizes the speedup comparison between our method with the baseline when training on the CIFAR-100, which shows gains above $1.229 \times$ (and $1.300 \times$ on average).
In addition, the speedup of our method is also observed in Fig.~\ref{fig:training_loss_cifar10} and Fig.~\ref{fig:training_loss_cifar100}, which show our model achieving faster loss values similar to the baseline model when training, respectively, on the CIFAR-10 and CIFAR-100 for all six DCNN architectures.

\begin{table*}[!ht]
\centering
\caption{
TL Time Analysis: Our Method Outperforms the Baseline (Based on
\cite{Mathieu2014}) 
in Training, with an average speedup of $1.316 \times$, and
Inference, with an average speedup of $1.321$, on CIFAR-10.
}
\label{tab:model_training_time_cifar10}
{
\begin{tabular}{|c|c|c|c|c|c|c|c|c|c|c|}
\hline \hline
Architecture &
  \begin{tabular}[c]{@{}c@{}}Batch\\ Size\end{tabular} &
  Method &
  \begin{tabular}[c]{@{}c@{}}Training\\ Loss\end{tabular} &
  \begin{tabular}[c]{@{}c@{}}Training\\ Accuracy\end{tabular} &
  \begin{tabular}[c]{@{}c@{}}Validation\\ Loss\end{tabular} &
  \begin{tabular}[c]{@{}c@{}}Validation\\ Accuracy\end{tabular} &
  \begin{tabular}[c]{@{}c@{}}Duration\\ Training\\ (sec)\end{tabular} &
  \begin{tabular}[c]{@{}c@{}}Duration \\ Validation\\  (sec)\end{tabular} &
  \begin{tabular}[c]{@{}c@{}}Training \\ Speedup\\  ($T_b/T_m$)\end{tabular} &
  \begin{tabular}[c]{@{}c@{}}Validation \\ Speedup\\  ($T_b/T_m$)\end{tabular} \\ \hline \hline
VGG-16         & 4  & Baseline   & 0.07705 & 97.125 & 0.23592 & 93.360 & 87371 & 6409 & 1.000 & 1.000 \\ \hline \rowcolor{green!10}
VGG-16         & 4  & Our Method & 0.07173 & 97.764 & 0.23491 & 93.640 & 69441 & 5087 & 1.258 & 1.260 \\ \hline
DenseNet-121   & 8  & Baseline   & 0.07874 & 97.691 & 0.10685 & 96.430 & 67381 & 4804 & 1.000 & 1.000 \\ \hline \rowcolor{green!10}
DenseNet-121   & 8  & Our Method & 0.07890 & 98.010 & 0.10725 & 96.390 & 50714 & 3589 & 1.329 & 1.338 \\ \hline
EfficientNetB3 & 8  & Baseline   & 0.12404 & 96.099 & 0.07638 & 97.530 & 76117 & 4828 & 1.000 & 1.000 \\ \hline \rowcolor{green!10}
EfficientNetB3 & 8  & Our Method & 0.12398 & 96.099 & 0.07632 & 97.540 & 59200 & 3777 & 1.286 & 1.278 \\ \hline
Inception-V3   & 8  & Baseline   & 0.11788 & 96.099 & 0.12478 & 95.960 & 74980 & 4469 & 1.000 & 1.000 \\ \hline \rowcolor{green!10}
Inception-V3   & 8  & Our Method & 0.11881 & 95.860 & 0.12713 & 95.880 & 58435 & 3441 & 1.283 & 1.299 \\ \hline
AlexNet        & 64 & Baseline   & 0.02280 & 99.297 & 0.32768 & 90.640 & 2270  & 157  & 1.000 & 1.000 \\ \hline \rowcolor{green!10}
AlexNet        & 64 & Our Method & 0.02271 & 99.297 & 0.32763 & 90.610 & 1666  & 115  & 1.363 & 1.362 \\ \hline
ResNet-18      & 64 & Baseline   & 0.02630 & 99.219 & 0.14690 & 95.050 & 7612  & 523  & 1.000 & 1.000 \\ \hline \rowcolor{green!10}
ResNet-18      & 64 & Our Method & 0.02627 & 99.297 & 0.14701 & 95.050 & 5534  & 376  & 1.376 & 1.390 \\ \hline \hline 
\end{tabular}
}
\end{table*}
\begin{table*}[ht]
\centering
\caption{TL Time Analysis: Our Method Outperforms the Baseline (Based on
\cite{Mathieu2014}) 
in Training, with an average speedup of $1.299 \times$, and
Inference, with an average speedup of $1.300$, on CIFAR-100.}
\label{tab:model_training_time_cifar100}
{
\begin{tabular}{|c|c|c|c|c|c|c|c|c|c|c|}
\hline \hline
Architecture &
  \begin{tabular}[c]{@{}c@{}}Batch\\ Size\end{tabular} &
  Method &
  \begin{tabular}[c]{@{}c@{}}Training\\ Loss\end{tabular} &
  \begin{tabular}[c]{@{}c@{}}Training\\ Accuracy\end{tabular} &
  \begin{tabular}[c]{@{}c@{}}Validation\\ Loss\end{tabular} &
  \begin{tabular}[c]{@{}c@{}}Validation\\ Accuracy\end{tabular} &
  \begin{tabular}[c]{@{}c@{}}Duration\\ Training\\ (sec)\end{tabular} &
  \begin{tabular}[c]{@{}c@{}}Duration \\ Validation\\  (sec)\end{tabular} &
  \begin{tabular}[c]{@{}c@{}}Training \\ Speedup\\  ($T_b/T_m$)\end{tabular} &
  \begin{tabular}[c]{@{}c@{}}Validation \\ Speedup\\  ($T_b/T_m$)\end{tabular} \\ \hline \hline
VGG-16         & 4  & Baseline   & 0.33055 & 89.776 & 1.06742 & 74.390 & 87430 & 6415 & 1.000 & 1.000 \\ \hline \rowcolor{green!10}
VGG-16         & 4  & Our Method & 0.33410 & 89.297 & 1.06839 & 74.230 & 69501 & 5092 & 1.258 & 1.260 \\ \hline 
DenseNet-121   & 4  & Baseline   & 0.89766 & 76.190 & 0.65382 & 80.880 & 74002 & 5263 & 1.000 & 1.000 \\ \hline \rowcolor{green!10}
DenseNet-121   & 4  & Our Method & 0.90051 & 75.000 & 0.65379 & 80.810 & 60225 & 4359 & 1.229 & 1.207 \\ \hline 
EfficientNetB3 & 8  & Baseline   & 0.59506 & 83.135 & 0.46725 & 85.840 & 76172 & 4828 & 1.000 & 1.000 \\ \hline \rowcolor{green!10}
EfficientNetB3 & 8  & Our Method & 0.59506 & 83.135 & 0.46725 & 85.840 & 59235 & 3778 & 1.286 & 1.278 \\ \hline 
InceptionV3    & 8  & Baseline   & 0.66419 & 79.379 & 0.88544 & 76.170 & 74894 & 4462 & 1.000 & 1.000 \\ \hline \rowcolor{green!10}
InceptionV3    & 8  & Our Method & 0.66513 & 79.379 & 0.88508 & 76.210 & 58369 & 3435 & 1.283 & 1.299 \\ \hline 
AlexNet        & 64 & Baseline   & 0.08057 & 97.656 & 1.23009 & 70.220 & 2271  & 158  & 1.000 & 1.000 \\ \hline \rowcolor{green!10}
AlexNet        & 64 & Our Method & 0.08076 & 97.656 & 1.22947 & 70.280 & 1670  & 115  & 1.360 & 1.369 \\ \hline 
ResNet-18      & 64 & Baseline   & 0.21980 & 95.000 & 0.64454 & 80.660 & 7617  & 525  & 1.000 & 1.000 \\ \hline \rowcolor{green!10}
ResNet-18      & 64 & Our Method & 0.21983 & 95.000 & 0.64448 & 80.650 & 5538  & 378  & 1.375 & 1.387 \\ \hline \hline
\end{tabular}
}
\end{table*}

\subsection{Additional Resources}

The source code and output for the experimental results presented in this section are available online\footnote{https://github.com/eduardo4jesus/Phasor-driven}.
\section{Discussion}\label{sec:discussion}

Traditionally, FFT-based CNN requires $4$ real-value multiplication and $2$ real-value additions to process each element of the convolution layer output in the spectral domain.
Such cost is inherited from the complex multiplication cost of the rectangular form,  \eqref{eq:rect_product}, and it is approximated by $4 B f_2 f_1 N^2$ FLOPS in the original estimate of \eqref{eq:flops-baseline}.
Our method proposes using phasors to represent complex numbers as an alternative to the rectangular form.
In this representation, the complex multiplication can be computed with only $1$ real-value multiplication and $1$ real-value addition, \eqref{eq:phasor_product}, yielding an estimate of $B f_2 f_1 N^2$ FLOPS. 
Such reduction by $3/4$ in the number of FLOPS is reflected in the speedup presented in Tables~\ref{tab:model_batch_profiling_cifar10} to Table~\ref{tab:model_training_time_cifar100}, in which we observe that executions with larger batch sizes, $B$, tend to benefit more, which is expected given the FLOPS estimate.

In addition to Table~\ref{tab:model_training_time_cifar10} and Table~\ref{tab:model_training_time_cifar100},  which show the final model performance in terms of loss and accuracy, Fig.~\ref{fig:training_loss_cifar10} and Fig.~\ref{fig:training_loss_cifar100} show that our method has equivalent performance to the baseline method.
This is expected due to the nature of our method, which is mathematically equivalent to the baseline method, differing only due to the limitation of the numerical representation of float numbers.
Compared to the baseline, our approach requires two extra steps, as shown in Fig.~\ref{fig:methodology}: \texttt{Step\#2} for computing \eqref{eq:complex-abs} and \eqref{eq:complex-angle}; and \texttt{Step\#4} for calculating \eqref{eq:rectangular}.
Despite these additional steps, the experimental results show that our approach yields enough acceleration to compensate for those extra costs while significantly outperforming the baseline in all scenarios.

Note that Tables~\ref{tab:model_batch_profiling_cifar10} to Table~\ref{tab:model_training_time_cifar100} also shows that the higher the batch size, the larger the speedup gain, such characteristic is expected from \eqref{eq:flops-baseline}.
Though larger values are desirable, the batch size value is bound to the memory usage constraints.
This trade-off of memory usage and speedup is not particular to our method but inherited from the adopted baseline.

Furthermore, our method can be combined with other approaches in the literature that focus only on reducing the cost of the FFT operations to yield even higher gains.
Though the experimental results were obtained using GPU, our method is platform agnostic; hence, it can be easily translated to existing FFT-based CNNs used in embedded applications.

\section{Conclusion} \label{sec:conclusion}

This paper investigates a phasor-based computational method to accelerate the training and inference speed of FFT-based CNNs. 
The proposed method benefits from the lower number of operations required by the phasor representation to multiply the Fourier representations of the inputs and kernels.
The experimental analysis proves that our method outperforms the speed of the baseline method up to $1.376 \times$ during training and up to $1.390 \times$ during the inference while yielding similar accuracy.

Future research can be dedicated to further investigating the potential phasor representation with other types of implementation, such as using optimized CUDA kernels or targeting embedded platforms.

\bibliographystyle{IEEEtran}
\bibliography{IEEEabrv,paper2/references}

\begin{thebibliography}{10}
\providecommand{\url}[1]{#1}
\csname url@samestyle\endcsname
\providecommand{\newblock}{\relax}
\providecommand{\bibinfo}[2]{#2}
\providecommand{\BIBentrySTDinterwordspacing}{\spaceskip=0pt\relax}
\providecommand{\BIBentryALTinterwordstretchfactor}{4}
\providecommand{\BIBentryALTinterwordspacing}{\spaceskip=\fontdimen2\font plus
\BIBentryALTinterwordstretchfactor\fontdimen3\font minus
  \fontdimen4\font\relax}
\providecommand{\BIBforeignlanguage}[2]{{%
\expandafter\ifx\csname l@#1\endcsname\relax
\typeout{** WARNING: IEEEtran.bst: No hyphenation pattern has been}%
\typeout{** loaded for the language `#1'. Using the pattern for}%
\typeout{** the default language instead.}%
\else
\language=\csname l@#1\endcsname
\fi
#2}}
\providecommand{\BIBdecl}{\relax}
\BIBdecl

\bibitem{ImageNet2014}
O.~Russakovsky, J.~Deng, H.~Su, J.~Krause, S.~Satheesh, S.~Ma, Z.~Huang,
  A.~Karpathy, A.~Khosla, M.~S. Bernstein, A.~C. Berg, and L.~Fei{-}Fei,
  ``Imagenet large scale visual recognition challenge,'' \emph{CoRR}, vol.
  abs/1409.0575, 2014.

\bibitem{AlexNet2012}
A.~Krizhevsky, I.~Sutskever, and G.~E. Hinton, ``Imagenet classification with
  deep convolutional neural networks,'' in \emph{Advances in Neural Information
  Processing Systems 25: 26th Annual Conference on Neural Information
  Processing Systems 2012. Proceedings of a meeting held December 3-6, 2012,
  Lake Tahoe, Nevada, United States}, 2012, pp. 1106--1114.

\bibitem{Vgg2015}
K.~Simonyan and A.~Zisserman, ``Very deep convolutional networks for
  large-scale image recognition,'' in \emph{3rd International Conference on
  Learning Representations, {ICLR} 2015, San Diego, CA, USA, May 7-9, 2015,
  Conference Track Proceedings}, Y.~Bengio and Y.~LeCun, Eds., 2015.

\bibitem{InceptionV3_2016}
C.~Szegedy, V.~Vanhoucke, S.~Ioffe, J.~Shlens, and Z.~Wojna, ``Rethinking the
  inception architecture for computer vision,'' in \emph{2016 {IEEE} Conference
  on Computer Vision and Pattern Recognition, {CVPR} 2016, Las Vegas, NV, USA,
  June 27-30, 2016}, 2016, pp. 2818--2826.

\bibitem{ResNet2016}
K.~He, X.~Zhang, S.~Ren, and J.~Sun, ``Deep residual learning for image
  recognition,'' in \emph{2016 {IEEE} Conference on Computer Vision and Pattern
  Recognition, {CVPR} 2016, Las Vegas, NV, USA, June 27-30, 2016}, 2016, pp.
  770--778.

\bibitem{Zhu2018}
H.~Zhu, M.~Akrout, B.~Zheng, A.~Pelegris, A.~Jayarajan, A.~Phanishayee,
  B.~Schroeder, and G.~Pekhimenko, ``Benchmarking and analyzing deep neural
  network training,'' in \emph{2018 {IEEE} International Symposium on Workload
  Characterization, {IISWC} 2018, Raleigh, NC, USA, September 30 - October 2,
  2018}, 2018, pp. 88--100.

\bibitem{You2018}
Y.~You, Z.~Zhang, C.~Hsieh, J.~Demmel, and K.~Keutzer, ``Imagenet training in
  minutes,'' in \emph{Proceedings of the 47th International Conference on
  Parallel Processing, {ICPP} 2018, Eugene, OR, USA, August 13-16, 2018}, 2018,
  pp. 1:1--1:10.

\bibitem{Mikami2019}
H.~Mikami, H.~Suganuma, Y.~Tanaka, Y.~Kageyama \emph{et~al.}, ``Massively
  distributed sgd: Imagenet/resnet-50 training in a flash,'' \emph{arXiv
  preprint arXiv:1811.05233}, 2018.

\bibitem{Yamazaki2019}
M.~Yamazaki, A.~Kasagi, A.~Tabuchi, T.~Honda, M.~Miwa, N.~Fukumoto, T.~Tabaru,
  A.~Ike, and K.~Nakashima, ``Yet another accelerated {SGD:} resnet-50 training
  on imagenet in 74.7 seconds,'' \emph{CoRR}, vol. abs/1903.12650, 2019.

\bibitem{Zhang2019}
Q.~Zhang, M.~Zhang, T.~Chen, Z.~Sun, Y.~Ma, and B.~Yu, ``Recent advances in
  convolutional neural network acceleration,'' \emph{Neurocomputing}, vol. 323,
  pp. 37--51, 2019.

\bibitem{BLAS}
``{BLAS} (basic linear algebra subprograms),'' \url{https://netlib.org/blas/},
  accessed: 2024-02-05.

\bibitem{Lavin2016}
A.~Lavin and S.~Gray, ``Fast algorithms for convolutional neural networks,'' in
  \emph{2016 {IEEE} Conference on Computer Vision and Pattern Recognition,
  {CVPR} 2016, Las Vegas, NV, USA, June 27-30, 2016}, 2016, pp. 4013--4021.

\bibitem{Mathieu2014}
M.~Mathieu, M.~Henaff, and Y.~LeCun, ``Fast training of convolutional networks
  through ffts,'' in \emph{2nd International Conference on Learning
  Representations, {ICLR} 2014, Banff, AB, Canada, April 14-16, 2014,
  Conference Track Proceedings}, 2014.

\bibitem{Vasilache2015}
N.~Vasilache, J.~Johnson, M.~Mathieu, S.~Chintala, S.~Piantino, and Y.~LeCun,
  ``Fast convolutional nets with fbfft: {A} {GPU} performance evaluation,'' in
  \emph{3rd International Conference on Learning Representations, {ICLR} 2015,
  San Diego, CA, USA, May 7-9, 2015, Conference Track Proceedings}, 2015.

\bibitem{CuDNN}
\BIBentryALTinterwordspacing
S.~Chetlur, C.~Woolley, P.~Vandermersch, J.~Cohen, J.~Tran, B.~Catanzaro, and
  E.~Shelhamer, ``cudnn: Efficient primitives for deep learning,'' \emph{CoRR},
  vol. abs/1410.0759, 2014. [Online]. Available:
  \url{http://arxiv.org/abs/1410.0759}
\BIBentrySTDinterwordspacing

\bibitem{PyTorch2019}
A.~Paszke, S.~Gross, F.~Massa, A.~Lerer, J.~Bradbury, G.~Chanan, T.~Killeen,
  Z.~Lin, N.~Gimelshein, L.~Antiga, A.~Desmaison, A.~K{\"{o}}pf, E.~Z. Yang,
  Z.~DeVito, M.~Raison, A.~Tejani, S.~Chilamkurthy, B.~Steiner, L.~Fang,
  J.~Bai, and S.~Chintala, ``Pytorch: An imperative style, high-performance
  deep learning library,'' in \emph{Advances in Neural Information Processing
  Systems 32: Annual Conference on Neural Information Processing Systems 2019,
  NeurIPS 2019, December 8-14, 2019, Vancouver, BC, Canada}, 2019, pp.
  8024--8035.

\bibitem{FacebookCudaLibrary}
``Facebook's cuda library,'' \url{https://github.com/facebookarchive/fbcuda},
  accessed: 2024-02-05.

\bibitem{Highlander2015}
T.~Highlander and A.~Rodriguez, ``Very efficient training of convolutional
  neural networks using fast fourier transform and overlap-and-add,'' in
  \emph{Proceedings of the British Machine Vision Conference 2015, {BMVC} 2015,
  Swansea, UK, September 7-10, 2015}, 2015, pp. 160.1--160.9.

\bibitem{Abtahi2018}
T.~Abtahi, C.~Shea, A.~M. Kulkarni, and T.~Mohsenin, ``Accelerating
  convolutional neural network with {FFT} on embedded hardware,'' \emph{{IEEE}
  Trans. Very Large Scale Integr. Syst.}, vol.~26, no.~9, pp. 1737--1749, 2018.

\bibitem{Lin2019}
J.~Lin and Y.~Yao, ``A fast algorithm for convolutional neural networks using
  tile-based fast fourier transforms,'' \emph{Neural Process. Lett.}, vol.~50,
  no.~2, pp. 1951--1967, 2019.

\end{thebibliography}

\end{document}